\documentclass[letterpaper, 10 pt, conference]{ieeeconf}  

\IEEEoverridecommandlockouts                              

\usepackage{graphics} 
\usepackage{epsfig} 
\usepackage{mathptmx} 
\usepackage{times} 
\usepackage{amsmath} 
\usepackage{amssymb}  

\usepackage{multirow}
\usepackage{subfigure}
\usepackage{etoolbox,siunitx}
\usepackage{tabularx}
\usepackage{algorithmic}
\usepackage{algorithm}
\usepackage[pagebackref,breaklinks,colorlinks]{hyperref}
\usepackage{booktabs}
\usepackage{wrapfig}
\usepackage{color}

\newcommand\cbar[3][mybar]{\colorbox{#1}{\color{black}\framebox(#2,#3){}}}
\definecolor{mybar}{rgb}{1.0, 0.5, 0.2}
\definecolor{purple}{rgb}{0.5, 0.0, 0.5}
\definecolor{orange}{rgb}{1, 0.65, 0}
\definecolor{lightgreen}{rgb}{0.68, 1, 0.18}
\definecolor{darkgreen}{rgb}{0.09, 0.32, 0.24}
\definecolor{darkred}{rgb}{0.6, 0, 0}
\definecolor{brown}{rgb}{0.64, 0.16, 0.16}
\definecolor{forestgreen}{rgb}{0.133, 0.545, 0.133}
\definecolor{source}{rgb}{0.266, 0.445, 0.765}
\definecolor{target}{rgb}{0.925, 0.489, 0.191}

\title{\LARGE \bf
ConDA: Unsupervised Domain Adaptation for LiDAR Segmentation via Regularized Domain Concatenation
}

\author{Lingdong Kong$^{1,*}$, Niamul Quader$^{2}$, Venice Erin Liong$^{2}$
\thanks{$^{1}$Lingdong Kong is with CNRS@CREATE and the National University of Singapore. Email:
        {\tt\small lingdong@comp.nus.edu.sg}.}%
\thanks{$^{2}$Niamul Quader and Venice Erin Liong are with Motional, Singapore. Email:
        {\tt\small \{niamul.quader,venice.liong\}@motional.com}.}%
\thanks{$^{*}$Work done as an autonomous vehicle research intern at Motional.}
}

\begin{document}

\maketitle
\thispagestyle{empty}
\pagestyle{empty}

\begin{abstract}

Transferring knowledge learned from the labeled source domain to the raw target domain for unsupervised domain adaptation (UDA) is essential to the scalable deployment of autonomous driving systems. State-of-the-art methods in UDA often employ a key idea: utilizing joint supervision signals from both source and target domains for self-training. In this work, we improve and extend this aspect. We present \textbf{ConDA}, a concatenation-based domain adaptation framework for LiDAR segmentation that: 1) constructs an intermediate domain consisting of fine-grained interchange signals from both source and target domains without destabilizing the semantic coherency of objects and background around the ego-vehicle; and 2) utilizes the intermediate domain for self-training. To improve the network training on the source domain and self-training on the intermediate domain, we propose an anti-aliasing regularizer and an entropy aggregator to reduce the negative effect caused by the aliasing artifacts and noisy pseudo labels. Through extensive studies, we demonstrate that ConDA significantly outperforms prior arts in mitigating domain gaps.

\end{abstract}

\section{INTRODUCTION}
Large-scale annotated data are desirable as they often yield robust and generalizable models. However, annotating semantic labels for 3D data like LiDAR point clouds \cite{LiDAR} in autonomous driving \cite{nuScenes,Panoptic-nuScenes,SemanticKITTI,Lyft-Level5,Waymo-Open,Argoverse,A2D2} is extremely expensive \cite{LiDAR-Point-Cloud-Generator,Survey-LiDAR-Semantic-Segmentation,Idan}.
This motivates us to explore unsupervised domain adaptation (UDA) for transferring knowledge learned from one domain, \textit{e.g.}, Boston, to another domain, \textit{e.g.}, Singapore, for scalable \textit{cross-city} deployment. 

UDA aims to tackle scenarios where a model is trained on labeled data from a \textit{source} domain and unlabeled data from a different but related \textit{target} domain, with the goal of enabling the model to perform well during target test time. A common practice in UDA is to jointly learn from both domains \cite{Survey-UDA-Semantic-Segmentation}. Prior works fall into two lines: 1) implicit learning domain-invariant features with discriminators via adversarial training \cite{DANN,GAN,DANN-Journal}, and 2) self-training with joint supervision signals from both the source (\textit{w/} ground-truth) and target (\textit{w/} pseudo-labels generated by confidence thresholding) domains~\cite{Pseudo-Labeling,Self-Training}. Recent studies have shown that the latter often yields more robust models with less computation cost \cite{CBST,DA-SAC}. These works, however, learn separately from source and target batches, and thus lack learning fine-grained interactions of objects and background in-between domains. Implementing an intermediate domain that can facilitate interactions to reduce the domain gap is intuitive. The ground-truth signals from the source domain can construct a ``shortcut" for correcting false target predictions in the local vicinity since feature representations close to each other tend to have the same semantic label \cite{Smooth}. Additionally, the target ``pseudo supervisions" -- which often contain large amounts of ignored labels -- can serve as a strong consistency regularization \cite{Strong} for the source domain. This guided supervision and regularization may yield a more adaptable and robust feature learning. It is not easy, however, to directly mix domains via interpolation \cite{MixUp,Interpolation-Consistency} or superposition \cite{CutMix,CutOut} for semantic segmentation as such techniques can corrupt the semantic coherency \cite{Strong}.

\begin{figure*}[t]
    \begin{center}
    \includegraphics[width=0.99\textwidth]{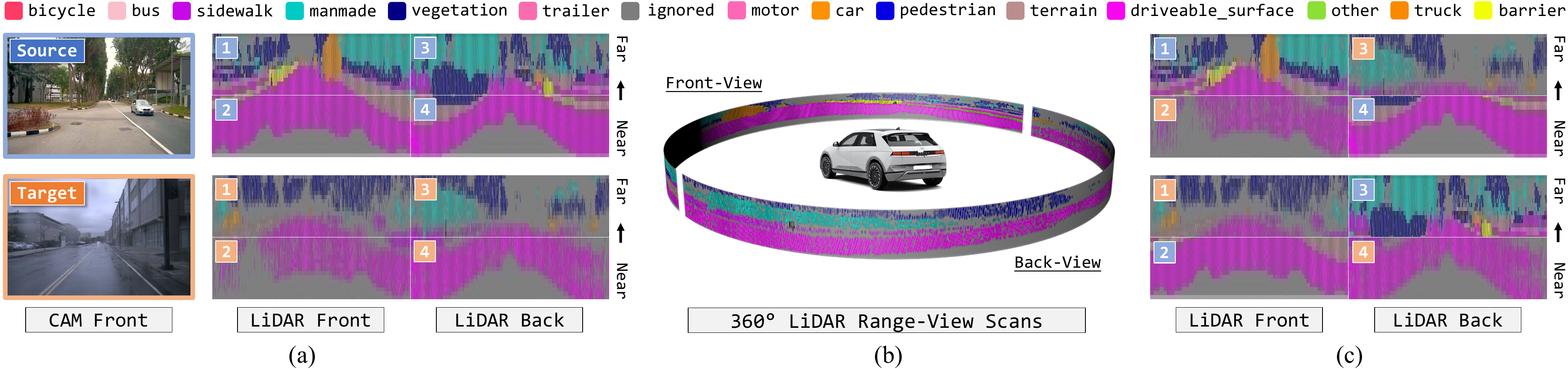}
    \end{center}
    \vspace{-0.52cm}
    \caption{Illustrative examples for domain concatenation. (a) Visual RGB and LiDAR range-view (RV) projections of the source (ground-truth) and target (pseudo-labels) domains. Images adopted from nuScenes \cite{nuScenes}. (b) Cylindrical representation of LiDAR RV. (c) Concatenated examples. Mixing domains using our ConDA strategy yields semantically realistic intermediate domain samples for self-training.}
    \label{fig:concat}
    \vspace{-0.3cm}
\end{figure*}

In this work, we propose a concatenation-based domain adaptation (ConDA) approach for cross-city UDA in LiDAR segmentation. ConDA enables the interactions of fine-grained semantic information in-between the source and target cities (domains) while not destabilizing the semantic coherency. We make the observations that although the overall distributions for the source and target domains are very different, there is a strong likelihood that similar objects and background tend to occupy particular regions in the LiDAR range-view (RV) around the ego-vehicle. We exploit this important correlation to construct the ConDA intermediate domain that selects non-overlapping regions of point clouds from both source and target domains and concatenates them together, while maintaining the relative positions from the ego-vehicle. As shown in Fig.~\ref{fig:concat}, concatenating different regions (\textit{e.g.}, front-top, front-bottom, back-top, and back-bottom) of RV stripes from different domains can still preserve semantic consistency. We will revisit this formally in Sec.~\ref{sec:domain-concatenation}.

The proposed domain concatenation strategy provides a better solution for model self-training via interchanged supervisions from both the source domain and the target domain. It is worth noting that the quality of the pseudo-labels -- and in turn the generalizability of the initial training on the source domain -- becomes essential since erroneous ``pseudo supervision" can do evil in self-training. We counter this problem from two perspectives as follows. First, the generalizability of the model can be improved by removing high-frequency noisy aliasing artifacts \cite{Impact-Anti-Aliasing,BlurPool} with careful placements of low-pass anti-aliasing filters \cite{DDAC}. However, empirically designing such filters is challenging in the context of UDA, since the lack of target domain ground-truth prevents any empirical experiments. To reduce the detrimental effect of high-frequency aliasing artifacts \cite{curriculum,spectralBias} without accessing target annotations, we propose a built-in regularization mechanism (Sec.~\ref{sec:anti-aliasing}) within each convolution block to regularize high-frequency representation learning during training, as well as to mitigate possible aliasing caused by region re-grouping. 
Furthermore, as the pseudo-labels are ``guessed" by the model trained on the source domain, it is intuitive to leverage the uncertainty \cite{Entropy-Semi-Supervised-Learning} of such ``guesses" for filtering non-confident selections. Entropy \cite{Entropy} -- a measure of choice freedom -- has been proven conducive for estimating prediction uncertainty \cite{vivo,Uncertainty-Aware-Pseudo-Label-Refinery}. Different from prior arts that implicitly minimize entropy in an adversarial way \cite{ADVENT,IntraDA}, we design an entropy aggregator (Sec.~\ref{sec:entropy-aggregator}) which explicitly eliminates high entropy target predictions and thus improves the overall quality for the intermediate domain supervisions.

Overall, this work has the following key contributions:
\begin{itemize}
    \item We propose ConDA, a novel framework that facilitates fine-grained interactive learning in-between domains. To the best of our knowledge, we are the first to explore cross-city UDA for uni-modal LiDAR segmentation, which can serve as a baseline for future research.
    \item We design two efficient regularization techniques to reduce detrimental aliasing artifacts and uncertain target predictions during model pre-training and self-training.
    \item We conduct comprehensive studies on the effects of our technical contributions on two challenging cross-city UDA scenarios. Our methods provide significant performance gains over state-of-the-art approaches.
\end{itemize}

\section{Related Work}
\noindent\textbf{UDA on Visual RGB}.
Adversarial training \cite{DANN,DANN-Journal} and self-training \cite{Pseudo-Labeling,Self-Training} dominate almost all kinds of 2D scene adaptation scenarios \cite{GTA5,SYNTHIA,Cityscapes,Day-Night}. Methods based on adversarial training adopt domain discriminators \cite{GAN,DCGAN} to implicitly search for domain-invariant features via distance measurements at different levels, \textit{i.e.}, input-level \cite{CyCADA,Photometric-Alignment,DACS,FDA,Reconstruction}, feature-level \cite{CLAN,DCAN,SSF-DAN}, and output-level \cite{AdaptSegNet,ADVENT,DADA}. However, such methods suffer from high computational costs and tend to be sensitive to hyperparameters and target domain changes \cite{MLCNet,xMUDA}. Self-training, on the other hand, offers a lighter option by jointly learning from both source and target domain supervisions \cite{SIM}, where the latter can be generated via confidence thresholding \cite{BDL,CBST,CRST}. Evidence shows that these methods -- albeit powerful in 2D -- become less effective in 3D \cite{xMUDA,DsCML}, which motivates us to design new methods w.r.t. the characteristics of the LiDAR data, such as the spatial consistency of the range-view representation. 

\noindent\textbf{UDA on LiDAR Data}. 
Adaptations are more challenging in 3D as point clouds are sparse, unstructured, and have limited visual cues compared to images \cite{Survey-DA-LiDAR,Sensor-Domain-Transfer,LiDARNet}. 
ePointDA \cite{ePointDA} learns a dropout noise rendering from real-world data to match synthetic data. Complete\&Label \cite{Complete-and-Label} targets on cross-sensor UDA where sparsity rather than city discrepancies serves as the major domain gap, which is beyond the scope of this work. xMUDA \cite{xMUDA} and DsCML \cite{DsCML} employ self-training alongside multi-modality learning for cross-city adaptations. However, the assumption of having access to synchronized RGB and LiDAR data in both source and target domains is not always practical, which limits such methods. Also, they use point/voxel backbones which require huge memory consumptions and suffer from low inference speed. Our framework is built upon mature 2D CNNs and does not require data from multiple modalities thus maintaining both simplicity and efficiency for practitioners.

\noindent\textbf{Domain Mixed Inputs}.
Mixing-based strategies \cite{MixUp,CutMix,CutOut} have been widely adopted in fully-~\cite{Mix3D,Domain-MixUp,Select-Label-Mix} and semi-~\cite{ClassMix,MixMatch,Interpolation-Consistency} supervised learning tasks, but very few touched UDA. SimROD \cite{SimROD} proposed to create 2x2 collages with two source and two target images for object detection. DACS \cite{DACS} cuts source objects out and pastes them onto the target images, alongside mixing corresponding source labels and target pseudo-labels. While the former \cite{SimROD} can corrupt the semantic consistency, the latter \cite{DACS} requires extra costs for the ``copy-paste" operation and lacks mixing background and target objects. Our ConDA provides more fine-grained interactions in-between domains and the concatenations are performed on the fly during training (via simple \texttt{torch.cat} or \texttt{tf.concat}) with almost zero cost. 

\begin{figure*}[!ht]
    \begin{center}
    \includegraphics[width=0.99\textwidth]{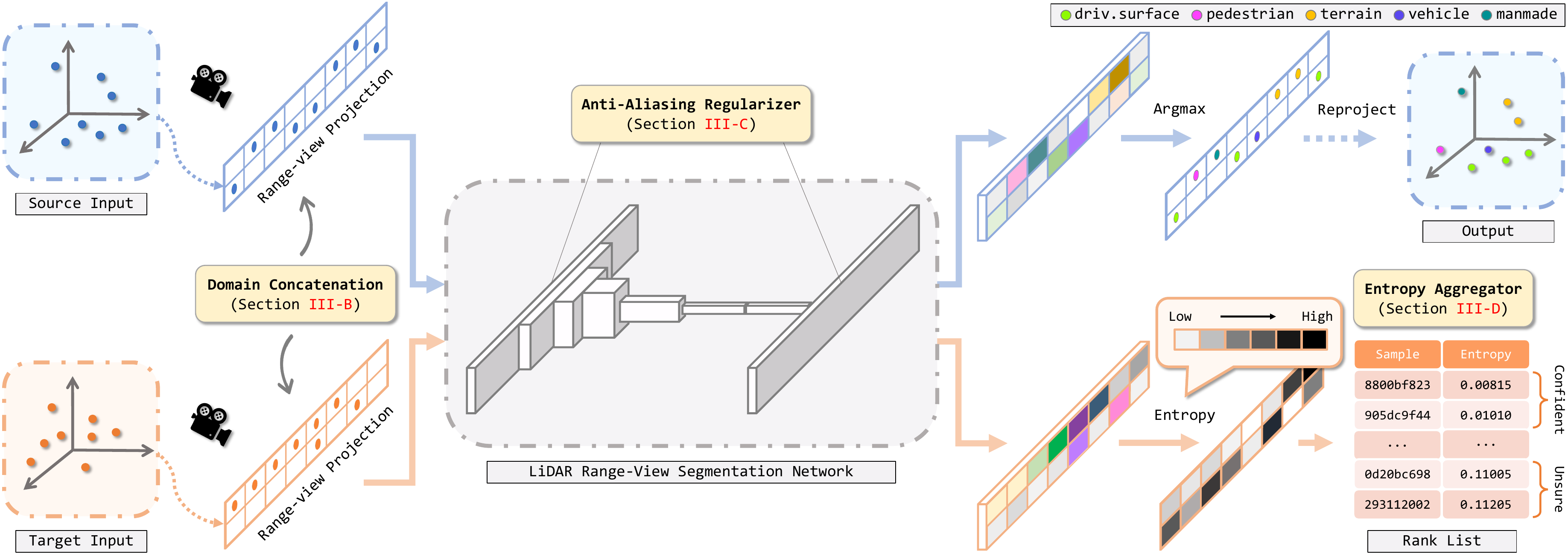}
    \end{center}
    \vspace{-0.5cm}
    \caption{Overview of our concatenation-based domain adaptation (ConDA) framework. After preprocessing (Sec.~\ref{sec:preliminaries}), sample stripes from both domains are mixed via RV concatenation (Sec.~\ref{sec:domain-concatenation}). The concatenated inputs are fed into the segmentation network for feature extraction. We include anti-aliasing regularizers inside convolution operations (Sec.~\ref{sec:anti-aliasing}) to suppress the learning of high-frequency aliasing artifacts. The segmented RV cells are then projected back to the point clouds. Here the target prediction part is omitted for simplicity. To mitigate the impediment caused by erroneous target predictions, we design an entropy aggregator (Sec.~\ref{sec:entropy-aggregator}) which splits samples into a confident set and an unsure set and disables the usage of samples from the latter set.}
    \label{fig:framework}
    \vspace{-0.2cm}
\end{figure*}

\noindent\textbf{Regularization}.
Anti-aliasing filters are popularly applied to reduce the aliasing artifacts in networks \cite{Impact-Anti-Aliasing,BlurPool,DDAC}. Such regularizers, however, have not been used in UDA. We conjecture that this is because the empirical observations required in filter design become impractical on the annotation-free target domain. In this work, we adopt a learning-based approach as an alternative to regularize aliasing artifacts. Another regularization is uncertainty estimation, which has been proved useful in semi-supervised learning \cite{Entropy-Semi-Supervised-Learning}. For UDA, IntraDA \cite{IntraDA} uses a discriminator to close the gap between the low-entropy and high-entropy samples. We design an entropy aggregator to disable the usage of non-confident pseudo-labels during domain adaptation, which does not rely on extra discriminator network and directly reduces the uncertainty for the intermediate domain supervisions.
\section{Technical Approach}

\subsection{Preliminaries}
\label{sec:preliminaries}
The proposed ConDA framework consists of three major components as shown in Fig.~\ref{fig:framework}.
Let ${A}$ denote a LiDAR point cloud. We project each point $\mathbf{o}=(x,y,z)$ on the 360$^{\circ}$ scan via a mapping $\Pi: \mathbb{R}^{3}\mapsto\mathbb{R}^{2}$ to a cylindrical RV image $\mathbf{a}\in\mathbb{R}^{6, h, w}$ with height $h$ and width $w$. $h$ is set based on the beam number of the sensor and $w$ is determined by the horizontal angular resolution. Each pixel in $\mathbf{a}$ consists of the point coordinates $(x,y,z)$, intensity, range $||\mathbf{o}||_{2}$, and a binary occupancy mask. Note that the RV projection preserves the range information and the spatial correspondence for the LiDAR scans, \textit{e.g.}, the near-front and far-front points are projected onto the left-bottom and left-top of the RV images, respectively, which is a unique feature of the LiDAR representations (Fig.~\ref{fig:concat}). To extract RV features efficiently, we design a seven-stage fully-convolutional network ${G}$ with strided convolutions \cite{FCN} and skip-connections \cite{ResNet} as our backbone. Since $h$ is much smaller than $w$, we only downsample the height at later stages. The segmentation head combines the upsampled outputs from the last four stages for multi-scale feature aggregations. Compared to previous RV networks \cite{SqueezeSegV3,RangeNet++,SalsaNext,3D-MiniNet}, our proposed $G$ offers a better trade-off between accuracy and speed, which is essential for LiDAR UDA.

\subsection{Domain Concatenation}
\label{sec:domain-concatenation}
In the context of UDA for LiDAR segmentation, we denote samples $\mathbf{a}$ from the labeled source domain and unlabeled target domain as ${A}_{s}=\{(\mathbf{a}_{s,m}, {q}_{s,m})\}_{m=1}^{M}$ and ${A}_{t}=\{(\mathbf{a}_{t,n})\}_{n=1}^{N}$, respectively, where $M$ and $N$ are the total number of source and target samples. For segmentation with $C$ classes, the network ${G}$ can be optimized in a supervised way with source samples by minimizing the cross-entropy loss as follows:
\vspace{-0.2cm}
\begin{equation}
\min_{\mathbf{w}}{L}_{s} = -\frac{1}{M}\sum_{m = 1}^{M}\sum_{c=1}^{C}{q}_{s,m}^{(c)} \log p(c|\mathbf{a}_{s,m},\mathbf{w}),
\label{equ:source_domain_loss}
\vspace{-0.05cm}
\end{equation}
where $\mathbf{w}$ denotes the weights of ${G}$; $p(\cdot)$ is the probability of class $c$ in the softmax output. 
Due to the lack of ground-truth in the target domain, we consider the target labels as hidden variables \cite{Pseudo-Labeling} and select the most confident target predictions on the existing model as one-hot ``pseudo-labels" $\hat{{q}}_{t}$. 
The learning objective for the target domain is:
\vspace{-0.2cm}
\begin{equation}
\begin{aligned}
\min_{\mathbf{w}}{L}_{t} = -\frac{1}{N}\sum_{n = 1}^{N}\sum_{c=1}^{C}\hat{{q}}_{t,n}^{(c)} \log p(c|\mathbf{a}_{t,n},\mathbf{w}),~~~~~~~~~\\
\text{s.t.} ~ \hat{{q}}_{t,n} =
\begin{cases}
\arg\max_{c}p(c|\mathbf{a}_{t,n},\mathbf{w}), &\text{if} \max(p(c|\mathbf{a}_{t,n},\mathbf{w}))\geq\theta\\
\text{ignored}, &\text{otherwise}
\end{cases}
\end{aligned}
\label{equ:target_domain_loss}
\end{equation}
where $\theta$ is a threshold for filtering non-confident pseudo-labels. Similar to \cite{CBST,CRST}, we set a proportion parameter $k$ to determine the class-wise thresholds $\theta_{c}$ for each class $c$ to balance the class distributions.
Now given a source batch $\{(\mathbf{a}_{s}, {q}_{s})\}$ and a target batch $\{(\mathbf{a}_{t}, {\hat{q}}_{t})\}$, our intermediate domain construction mechanism ${M}(\cdot)$ follows three steps. First, define a template for the total number of segregation regions along the near-far dimension ($m$) in the RV projections and around the ego-vehicle ($n$). Fig.~\ref{fig:concat}\textcolor{red}{a} shows an example of a domain concatenation with $m=2$ and $n=2$, having front-near, front-far, back-near, and back-far regions, respectively. Second, slice regions based on the above template for every sample in both batches. This gives $((b_s+b_t) \times m \times n)$ sliced stripes, where $b_{s}$ and $b_{t}$ are the batch sizes. Third, concatenate the stripes while keeping their spatial locations consistent, resulting in $(b_s+b_t)$ intermediate domain samples $\mathbf{a}_{\pi}$. Their labels $q_{\pi}$ can be obtained via the same arrangement of the original labels and pseudo-labels. The segmentation loss ${L}_{\pi}$ for this intermediate domain can be computed using the mixed batch $\{(\mathbf{a}_{\pi}, {q}_{\pi})\}$ in a way similar to Eq.~\ref{equ:source_domain_loss}. The overall objective for self-training is to minimize ${L} = {L}_{s}(\mathbf{w}, {q}_{s}) + \sigma\cdot{L}_{\pi}(\mathbf{w}, {q}_{\pi})$,
where $\sigma$ is a coefficient that controls the probability of accessing the intermediate domain. As shown in Fig.~\ref{fig:concat}\textcolor{red}{c}, this simple concatenation approach mixes objects and background from both domains, while still preserving the overall consistency. This stability in semantic coherence comes from the priors that the LiDAR point clouds are unstructured and extremely sparse even after RV projections. Take nuScenes~\cite{Panoptic-nuScenes} as an example. We find that on average $59.93\%$ of RV cells are empty, which could downplay the negative impact of region rearrangement since the degree of continuity is low. 

\subsection{Anti-Aliasing Regularizer}
\label{sec:anti-aliasing}
We formulate a regularizer that is built within each convolution filter in $G$ to reduce learning from aliasing artifacts. Our goal is to impose regularization on high-frequency representation learning since they are more susceptible to aliasing artifacts \cite{BlurPool,DDAC}. We achieve this via $\textbf{f}_{c,r} = \textbf{f}_r \odot \textbf{f}_c$, where $\textbf{f}_r$ denotes our regularizer which consists of learnable parameters having the same size as the convolution filter $\textbf{f}_c$; $\textbf{f}_{c,r}$ is the regularized filter kernel of each convolution; $\odot$ is the Hadamard multiplication. Note that during the earlier stages of training, the network tends to learn low-frequency representations \cite{spectralBias,curriculum} that are robust to aliasing artifacts. This will thereby update $\textbf{f}_r$ such that it becomes more suited for low-frequency representation learning of $\textbf{f}_{c,r}$. In later phases of training or UDA self-training, however, networks are more inclined to learn increasingly higher frequency representations and thus become more susceptible to aliasing artifacts \cite{Impact-Anti-Aliasing}. The modulation of our $\textbf{f}_r$ on $\textbf{f}_c$ regularizes gradient updates corresponding to these high-frequency representations and in particular, regularizes the ones that are considerably different from the earlier network learning. This implicit regularization mechanism at later stages of training makes $\textbf{f}_{c,r}$ more resistant to aliasing artifacts than the plain $\textbf{f}_c$. Notably, since $\textbf{f}_c$ and $\textbf{f}_r$ are both constants during inference, the regularized kernel $\textbf{f}_{c,r}$ only needs to be computed once at the end of training and can be used at inference without adding any additional computational cost or structural changes to the network -- this makes our regularizer unique among all previous anti-aliasing mechanisms.

\subsection{Entropy Aggregator}
\label{sec:entropy-aggregator}
Given the fact that the pseudo-labels generated from the source pre-trained model tend to be noisy \cite{Uncertainty-Aware-Pseudo-Label-Refinery}, we design an entropy aggregator to disable the access of non-confident target predictions and thus improve the overall quality of the intermediate domain supervisions. More formally, given a target sample $\mathbf{a}_{t,n}$, its entropy map ${E}_{n}$ composed of the normalized pixel-wise entropies can be calculated as follows:
\begin{equation}
\begin{aligned}
{E}_{n} = \frac{-1}{\log(C)}\sum^{C}_{c=1}p(c|\mathbf{a}_{t,n},\mathbf{w})\log p(c|\mathbf{a}_{t,n},\mathbf{w}).
\end{aligned}
\label{equ:entropy}
\end{equation}
The median value $\mathbf{e}_{n}$ of ${E}_{n}$ for the target sample $\mathbf{a}_{t,n}$ is used as the global-level indicator of uncertainty, which is more robust than the average value due to the large ``noisy" predictions for the empty RV cells. The target set ${A}_{t}=\{(\mathbf{a}_{t,n})\}_{n=1}^{N}$ is re-organized based on the entropy ranking and only the pseudo-labels from the top $\varpi$ most confident target samples are included as the supervisions for the intermediate domain.
\section{Experiments and Analysis}

\subsection{Settings}
\noindent\textbf{Data}. We construct two RV-based cross-city UDA scenarios with nuScenes~\cite{nuScenes,Panoptic-nuScenes} -- a large-scale autonomous driving database widely adopted in academia. We split samples based on their geographic locations. This gives 15695 and 12435 training samples and 3090 and 2929 evaluation samples for Boston and Singapore, respectively. All training samples are used as the source/target for adaptations. Different from xMUDA \cite{xMUDA} which only assigns semantic labels to points inside bounding boxes with 4 object and 1 background classes, we adopt the \textit{lidarseg} subset in nuScenes which contains 16 classes and fine-grained point-level annotations. 

\begin{table}
\caption{Effectiveness for each component in ConDA. Evaluated on the Boston $\rightarrow$ Singapore adaptation setting.}
\vspace{-0.3cm}
\centering\scalebox{0.89}{
\footnotesize
\setlength\fboxsep{0pt}
\begin{tabularx}
{\linewidth}{@{}>{\raggedleft\arraybackslash}m{27mm}|S[table-format=2.1]|X@{}}
\toprule
\multicolumn{1}{>{\centering\arraybackslash}m{18mm}|}{$\Delta$} & mIoU & Configuration \\
\midrule
-12.0 \cbar{56}{5} & 34.9 & No adaptation (source-only) \\
-6.2 \cbar{32.0}{5} & 40.7 & Baseline (vanilla self-training) \\
-5.0 \cbar{25}{5} & 41.9 & += Anti-aliasing filters (Sec.~\ref{sec:anti-aliasing}) \\
-1.6 \cbar{15}{5} & 45.3 & += Domain concatenation (Sec.~\ref{sec:domain-concatenation}) \\
-1.0 \cbar{9}{5} & 45.9 & += Entropy aggregator (Sec.~\ref{sec:entropy-aggregator}) \\
\toprule
0.0 & 46.9 & Full ConDA framework \\
\bottomrule
\end{tabularx}}
\vspace{-0.15cm}
\label{tab:ablation}
\end{table}

\begin{figure}[t]
    \begin{center}
    \includegraphics[width=0.44\textwidth]{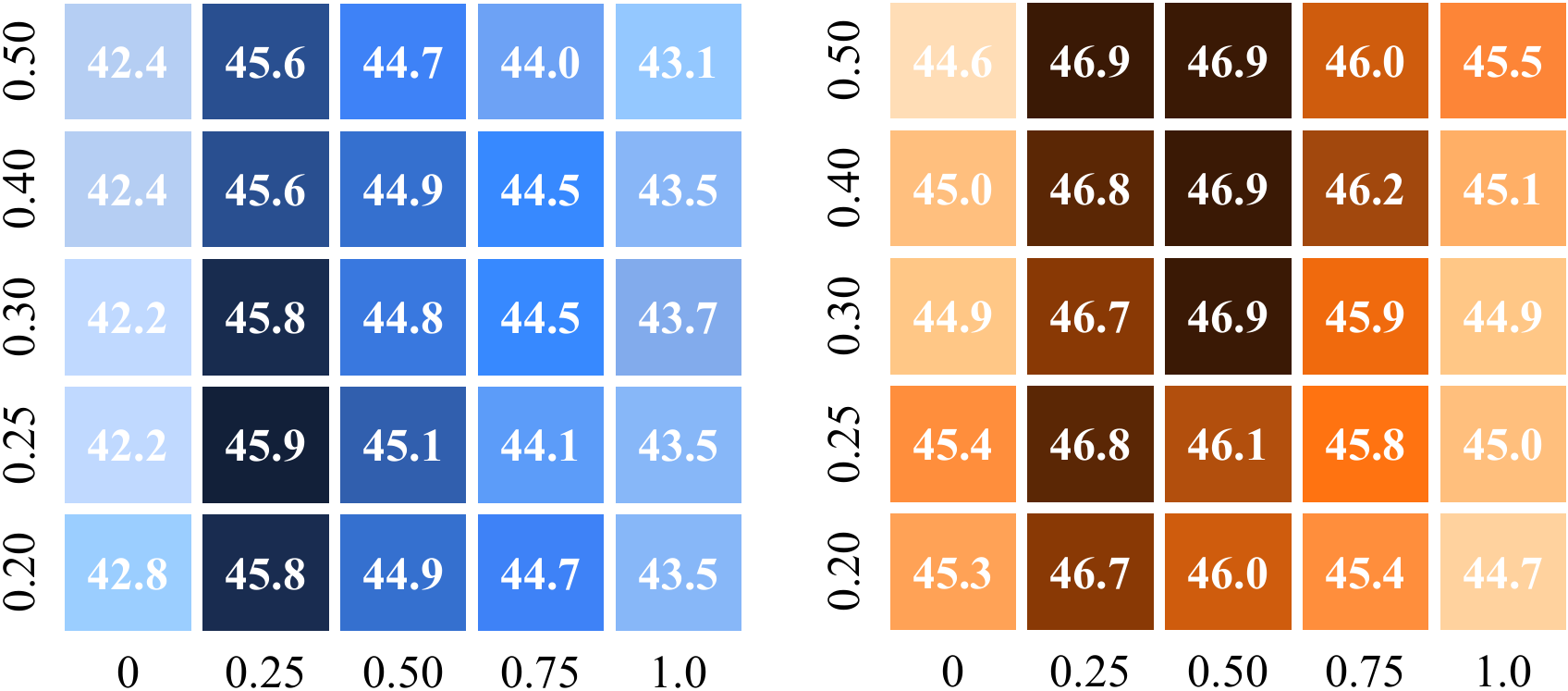}
    \end{center}
    \vspace{-0.52cm}
    \caption{Sensitivity analysis for the concatenation probability $\sigma$ (horizontal axis) and proportion threshold $k$ (vertical axis) in self-training round $1$ (left) and round $2$ (right). The darker the color, the higher the mIoU score.}
    \label{fig:ablation-params}
    \vspace{-0.35cm}
\end{figure}

\begin{table*}[!ht]
\caption{Adaptation results for different RV concatenation strategies (\textit{cf.}~Fig.~\ref{fig:concat-examples}) and other mixing techniques.}
\vspace{-0.3cm}
\centering\scalebox{0.77}{
\begin{tabular}{c|c|ccccccccccccccc|ccccc}
\toprule
Method & ~w/o~ & ~a~ & ~b~ & ~c~ & ~d~ & ~e~ & ~f~ & ~g~ & ~h~ & ~i~ & ~j~ & ~k~ & ~l~ & ~m~ & ~n~ & ~o~ & ~\cite{SimROD}~ & ~\cite{MixUp}~ & ~\cite{CutOut}~ & ~\cite{CutMix}~ & ~\cite{Mix3D}~
\\\midrule
mIoU & $41.9$ & $43.4$ & $43.5$ & $43.7$ & $44.5$ & $44.6$ & $43.9$ & $44.5$ & $44.6$ & $\mathbf{45.3}$ & $45.0$ & $45.2$ & $\mathbf{45.3}$ & $44.9$ & $44.6$ & $44.2$ & $33.7$ & $41.1$ & $43.1$ & $42.6$ & $40.1$
\\\bottomrule
\end{tabular}}
\vspace{-0.3cm}
\label{tab:concat-examples}
\end{table*}

\noindent\textbf{Implementation Details}. We project point clouds into RV images of size $32\times1920$ as the inputs for $G$ (\textit{cf.}~Sec.~\ref{sec:preliminaries}). It is first trained from scratch with only source samples for $80$ epochs and then fine-tuned under our entropy aggregator-guided self-training procedure. Both rounds are trained for $20$ epochs. We denote results for the supervised learning and direct adaptation as ``oracle" and ``source-only". Since this is a new benchmark, we could only compare ConDA with state-of-the-art adversarial \cite{AdaptSegNet,ADVENT,DADA}, self-training \cite{CBST,BDL,IntraDA}, and consistency training \cite{MeanTeacher,CPS} methods originally tested for 2D adaptations. For fairness, we replace their backbones with our RV network and keep other configurations in default. 
For methods based on self-training, we generate their pseudo-labels offline as in \cite{BDL,CBST}. All methods are implemented using PyTorch on NVIDIA Tesla V100 GPUs. 

\noindent\textbf{Evaluation Metrics}. We follow the conventional reporting of the intersection-over-union (IoU) scores ($\%$) over each class,  the mean IoU (mIoU) and the frequency-weighted IoU (FIoU) scores ($\%$) over all classes in our experiments.

\begin{figure}[t]
    \begin{center}
    \includegraphics[width=0.44\textwidth]{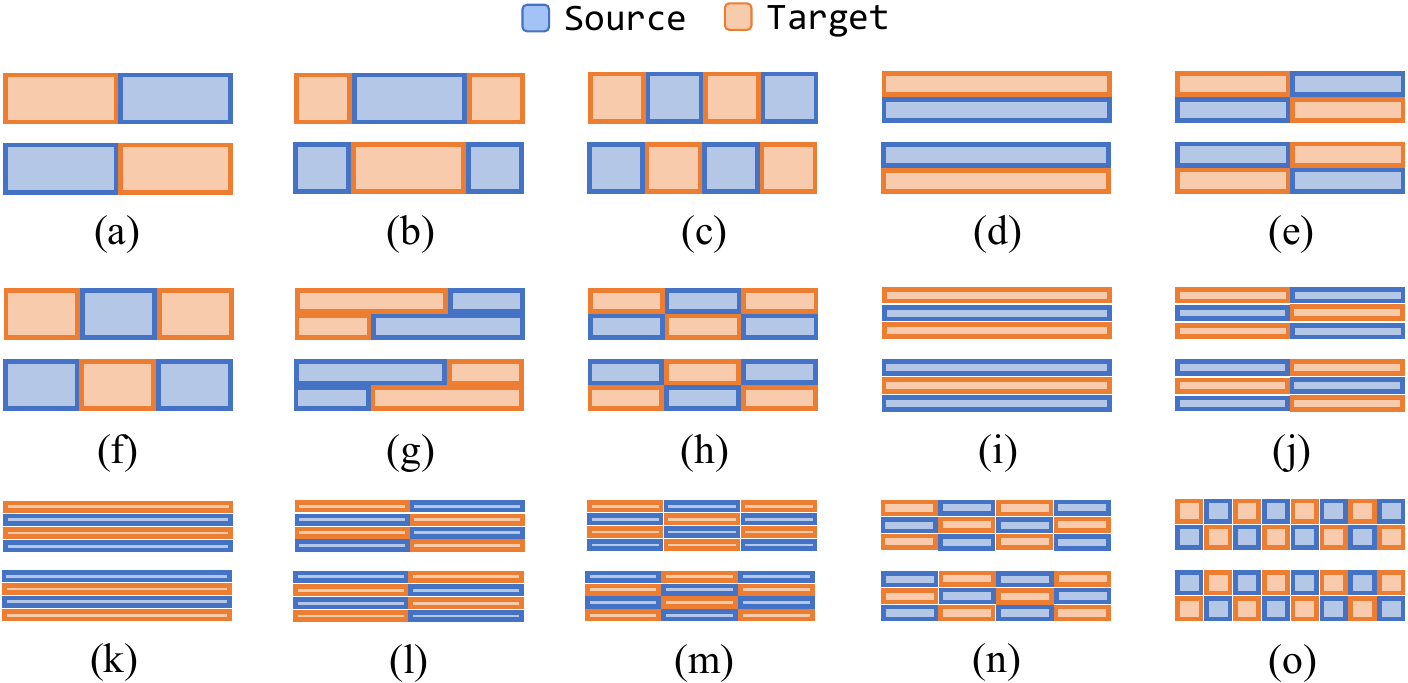}
    \end{center}
    \vspace{-0.5cm}
    \caption{ConDA RV concatenation strategies with various granularities.}
    \label{fig:concat-examples}
    \vspace{-0.4cm}
\end{figure}

\subsection{Ablation Studies}

\noindent\textbf{Q1: \textit{What is the effect for each component in ConDA?}}

\noindent\textbf{A1:} We adopt the Boston $\rightarrow$ Singapore setting in our ablation studies without the loss of generality. We stratify the three major components in our framework and show their impacts in Tab.~\ref{tab:ablation}. Specifically, the anti-aliasing regularizer offers an improvement of $1.2\%$ mIoU over the baseline and surpasses the source-only case by $7.0\%$ mIoU. Comparing the frequency spectrum of 3x3 kernels from the network, we also find that the ratio of \textit{`average low-frequency amplitudes'} to \textit{`average high-frequency amplitudes'} is 23.2$\%$ higher (statistically significant with t-test: $p<0.001$) for the network trained with our regularizer. On top of that, our domain concatenation further improves $3.4\%$ mIoU. Another boost of $1.6\%$ mIoU is achieved under the two-round guidance self-training of our entropy aggregator. Overall, our framework significantly improves the adaptation results from $34.9\%$ mIoU to $46.9\%$ mIoU, which corresponds to nearly a $34.4\%$ relative improvement over the source-only.

\noindent\textbf{Q2: \textit{What are the optimal hyperparameters for ConDA?}}

\noindent\textbf{A2:} 
We conduct extensive experiments to show the best possible selections for the hyperparameters. Specifically, the vertical and horizontal axes of Fig.~\textcolor{red}{3} show the impact of the proportion parameter $k$ and the concatenation parameter $\sigma$ during the two-round self-training. We find that lower values for $k$ (\textit{e.g.} $0.25$) in round $1$ and relatively higher values (\textit{e.g.} $0.5$) in round $2$ tend to give higher mIoU. We conjecture that this relatively conservative choice (or lower value) at round $1$ is helpful since pseudo-labels tend to be noisy at this stage. The quality of pseudo-label gets much better at round $2$ and including more of them gives a positive impact on the performance.  As for $\sigma$, we observe that the best possible scores are achieved between $0.25$ and $0.50$. Training \textit{w/o} ($\sigma=0$) or \textit{w/} all ($\sigma=1$) concatenated samples does not perform well. For $\varpi$ (after setting both $k$ and $\sigma$ as $0.25$ without the loss of generality), we find that large $\varpi$ involves more false positives while small $\varpi$ limits the diversity. A compromise value like $0.50$ gives the best possible scores.

\begin{figure}[t]
    \begin{center}
    \includegraphics[width=0.49\textwidth]{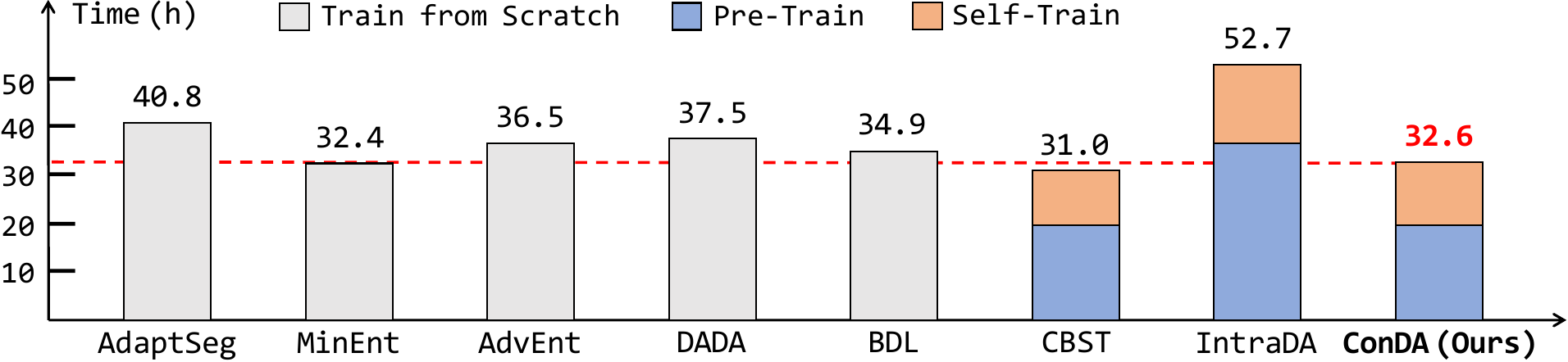}
    \end{center}
    \vspace{-0.5cm}
    \caption{Training time (in hours) needed for each method during adaptations.}
    \label{fig:train_time}
    \vspace{-0.3cm}
\end{figure}

\noindent\textbf{Q3: \textit{What is the best practice for RV concatenation?}}

\noindent\textbf{A3:} Besides the intuitive front-back RV concatenation, we also consider other scenarios as in Fig.~\ref{fig:concat-examples} and show their results in Tab.~\ref{tab:concat-examples}. As shown, strategies $i$ and $l$ perform the best while $j$ and $k$ offer competitive results. We note that: 1) increasing the granularity of the interactions between source and target tends to improve performance (strategies $a$ to $l$); 2) increasing the granularity beyond a certain limit (strategies $m$, $n$, and $o$) can deteriorate performance, which is likely due to the instability in semantic coherence of the objects and background in the concatenated stripes; and 3) fine-grained interactions along the vertical axes, \textit{i.e.,} near to far regions, perform better than interactions along the horizontal axes, \textit{i.e.,} bearing around the ego-vehicle (strategies $i$ and $f$), suggesting that the former likely yields better domain interactions while better maintaining the semantic consistency.

\begin{figure*}[t]
    \begin{center}
    \includegraphics[width=0.93\textwidth]{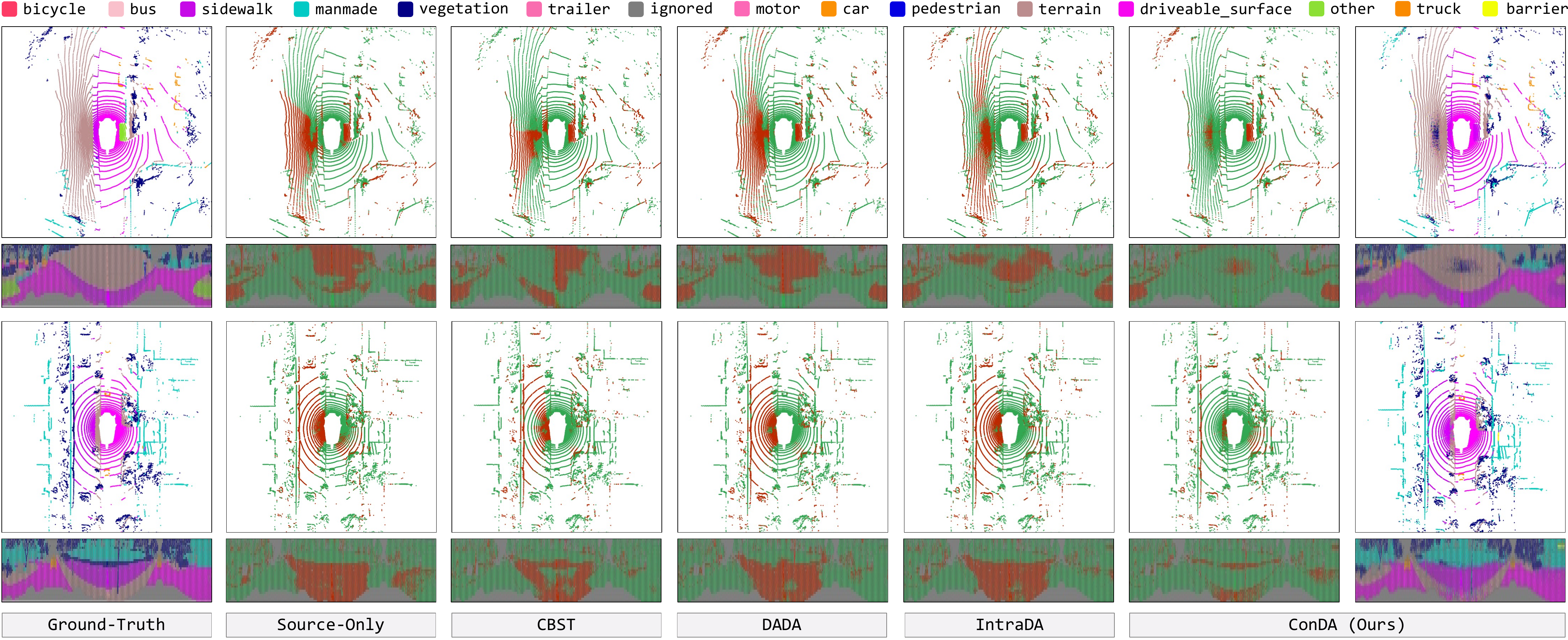}
    \end{center}
    \vspace{-0.52cm}
    \caption{Qualitative results from both the bird's eye view and range-view. To highlight the difference between the predictions and ground-truth, the \textcolor{forestgreen}{\textbf{correct}} and \textcolor{red}{\textbf{incorrect}} points/pixels are painted in \textcolor{forestgreen}{\textbf{green}} and \textcolor{red}{\textbf{red}}, respectively. Best viewed in color and zoom-ed in for details.}
    \label{fig:qualitative}
    \vspace{-0.05cm}
\end{figure*}

\begin{table*}[t]
\caption{Adaptation results for \textbf{Boston $\rightarrow$ Singapore}. The methods are grouped, from top to bottom, as adversarial training, self-training, both, and ours. All IoU scores are given in percentage ($\%$). Best score for each class is highlighted in \textbf{bold}.}
\vspace{-0.3cm}
\centering\scalebox{0.717}{
\begin{tabular}{r|cccccccccccccccc|cc}
\toprule
~Method~ & ~~\rotatebox{0}{barr}~ & ~\rotatebox{0}{bicy}~ & ~\rotatebox{0}{bus}~ & ~\rotatebox{0}{car}~ & ~\rotatebox{0}{const}~ & ~\rotatebox{0}{moto}~ & ~\rotatebox{0}{ped}~ & ~\rotatebox{0}{cone}~ & ~\rotatebox{0}{trail}~ & ~\rotatebox{0}{truck}~ & ~\rotatebox{0}{driv}~ & ~\rotatebox{0}{othe}~ & ~\rotatebox{0}{walk}~ & ~\rotatebox{0}{terr}~ & ~\rotatebox{00}{manm}~ & ~\rotatebox{0}{veg}~~ & ~\rotatebox{0}{FIoU}~$\uparrow$ & \rotatebox{0}{mIoU}~$\uparrow$
\\\midrule
Oracle~ & 79.5 & 33.6 & 87.5 & 88.9 & 37.6 & 75.6 & 70.5 & 50.3 & 0.0 & 76.2 & 95.1 & 53.7 & 60.2 & 74.4 & 83.4 & 85.5 & 84.2 & 65.8~
\\
Source-only~ & 29.3 & 1.3 & 52.0 & 71.4 & 7.2 & 11.7 & 42.6 & 12.2 & 0.0 & 30.4 & 85.9 & 12.7 & 32.6 & 41.0 & 62.5 & 65.9 & 64.3 & 34.9~
\\\midrule
AdaptSeg \cite{AdaptSegNet}~ & 28.0 & \textbf{7.2} & 60.9 & 70.7 & 7.7 & 17.4 & 45.5 & 14.3 & 0.0 & 36.4 & 88.1 & 28.4 & 36.0 & 43.1 & 63.0 & 66.7 & 66.1 & 38.3~
\\
MinEnt \cite{ADVENT}~ & 31.7 & 4.0 & 63.7 & 70.6 & 5.8 & 15.9 & 47.7 & 13.7 & \textbf{0.1} & 34.9 & 87.9 & 22.4 & 37.5 & 41.5 & 59.9 & 62.2 & 64.3 & 37.5~
\\
AdvEnt \cite{ADVENT}~ & 28.7 & 5.9 & 59.4 & 76.4 & 7.2 & 18.2 & 50.6 & \textbf{16.7} & 0.0 & 32.6 & 87.0 & 28.1 & 36.6 & 44.0 & 63.9 & 67.1 & 66.3 & 38.9~
\\
DADA \cite{DADA}~ & 27.4 & 4.9 & 60.0 & 67.7 & 7.3 & 15.9 & 44.4 & 14.7 & 0.0 & 33.9 & 87.1 & 21.2 & 34.9 & 42.1 & 62.2 & 64.9 & 64.8 & 36.8~
\\\midrule
BDL$_{\text{PL}}$ \cite{BDL}~ & 39.2 & 0.3 & 53.0 & 73.2 & 6.8 & 16.0 & 40.2 & 8.5 & 0.0 & 29.8 & 88.7 & 21.3 & 39.7 & 48.5 & 67.1 & 67.9 & 68.3 & 37.5~
\\
CBST \cite{CBST}~ & 39.4 & 5.3 & 66.1 & 75.6 & 9.3 & 20.7 & 47.8 & 14.9 & 0.0 & 34.1 & 88.4 & 25.5 & 38.1 & 49.9 & 66.7 & 68.5 & 68.6 & 40.7~
\\\midrule
AdaptSeg$_{\text{PL}}$ \cite{AdaptSegNet}~ & 29.9 & 0.3 & 47.9 & 64.4 & 4.9 & 7.4 & 28.4 & 4.6 & 0.0 & 24.8 & 83.1 & 21.8 & 38.3 & 46.5 & 67.1 & 68.9 & 66.0 & 33.7~
\\
IntraDA \cite{IntraDA}~ & 28.0 & 5.6 & 57.8 & 76.1 & 6.2 & 18.6 & 47.4 & 13.8 & 0.0 & 32.1 & 87.3 & 27.6 & 37.0 & 44.4 & 63.4 & 66.5 & 66.2 & 38.3~
\\\midrule
\textbf{ConDA~(Ours)}~ & \textbf{54.1} & 6.8 & \textbf{67.4} & \textbf{77.2} & \textbf{12.1} & \textbf{38.7} & \textbf{51.8} & 16.0 & 0.0 & \textbf{44.0} & \textbf{90.4} & \textbf{38.7} & \textbf{44.0} & \textbf{62.9} & \textbf{70.7} & \textbf{75.0} & \textbf{74.1} & \textbf{46.9}~
\\\bottomrule
\end{tabular}}
\vspace{-0.08cm}
\label{tab:bs2sg}
\end{table*}

\begin{table*}[t]
\caption{Adaptation results for \textbf{Singapore $\rightarrow$ Boston}. The methods are grouped, from top to bottom, as adversarial training, self-training, both, and ours. All IoU scores are given in percentage ($\%$). Best score for each class is highlighted in \textbf{bold}.}
\vspace{-0.3cm}
\centering\scalebox{0.715}{
\begin{tabular}{r|cccccccccccccccc|cc}
\toprule
~Method~ & ~~\rotatebox{0}{barr}~ & ~\rotatebox{0}{bicy}~ & ~\rotatebox{0}{bus}~ & ~\rotatebox{0}{car}~ & ~\rotatebox{0}{const}~ & ~\rotatebox{0}{moto}~ & ~\rotatebox{0}{ped}~ & ~\rotatebox{0}{cone}~ & ~\rotatebox{0}{trail}~ & ~\rotatebox{0}{truck}~ & ~\rotatebox{0}{driv}~ & ~\rotatebox{0}{othe}~ & ~\rotatebox{0}{walk}~ & ~\rotatebox{0}{terr}~ & ~\rotatebox{00}{manm}~ & ~\rotatebox{0}{veg}~~ & ~\rotatebox{0}{FIoU}~$\uparrow$ & \rotatebox{0}{mIoU}~$\uparrow$
\\\midrule
Oracle~ & 71.3 & 35.5 & 71.5 & 86.9 & 41.6 & 35.4 & 69.7 & 61.2 & 57.6 & 68.0 & 95.9 & 70.7 & 79.7 & 58.7 & 89.9 & 83.9 & 88.5 & 67.3~
\\
Source-only~ & 15.5 & 7.9 & 20.6 & 70.5 & 16.1 & 3.6 & 41.9 & 11.4 & 0.5 & 40.6 & 90.2 & 10.7 & 41.7 & 19.1 & 77.4 & 74.5 & 73.4 & 33.9~
\\\midrule
AdaptSeg \cite{AdaptSegNet}~ & 15.9 & 2.4 & 40.4 & 73.9 & 15.2 & 5.5 & 48.3 & 8.3 & 0.4 & 46.3 & 92.2 & 18.7 & 54.5 & 19.0 & 79.0 & 70.9~ & 76.2 & 36.9~
\\
MinEnt \cite{ADVENT}~ & \textbf{19.2} & 0.2 & 36.1 & 73.2 & 15.7 & 6.2 & 50.3 & 10.8 & 0.8 & 45.0 & 91.5 & 24.1 & 54.8 & 21.7 & 78.7 & 71.8 & 76.0 & 37.5~
\\
AdvEnt \cite{ADVENT}~ & 12.5 & \textbf{9.0} & 43.0 & 74.1 & 14.7 & 7.0 & 51.4 & 12.7 & 0.5 & 47.0 & 91.4 & 14.5 & 53.6 & 19.2 & 80.1 & 73.4 & 76.2 & 37.8~
\\
DADA \cite{DADA}~ & 18.5 & 2.9 & 35.5 & 73.0 & 15.0 & 6.5 & 49.3 & 11.0 & 1.6 & 43.6 & 91.8 & 12.2 & 52.7 & 19.7 & 79.8 & 73.4 & 76.1 & 36.7~
\\\midrule
BDL$_{\text{PL}}$ \cite{BDL}~ & 18.9 & 2.7 & 30.8 & 75.8 & 13.3 & 3.8 & 45.4 & 7.4 & 1.8 & 45.4 & 92.8 & 19.7 & 58.4 & 18.7 & 80.1 & 76.3 & 77.6 & 37.0~
\\
CBST \cite{CBST}~ & 17.7 & 1.4 & 33.6 & 75.0 & 13.3 & 6.4 & 52.3 & 12.3 & \textbf{1.9} & 46.9 & 92.5 & 22.8 & 57.2 & 19.7 & 80.2 & 77.3 & 77.5 & 38.1~
\\\midrule
AdaptSeg$_{\text{PL}}$ \cite{AdaptSegNet}~ & 10.5 & 0.6 & 33.5 & 71.3 & \textbf{17.2} & 5.2 & 41.9 & 11.4 & 1.0 & 43.5 & 90.4 & 18.5 & \textbf{60.1} & 20.0 & 80.0 & 74.6 & 76.1 & 36.3~
\\
IntraDA \cite{IntraDA}~ & 12.3 & 6.2 & 41.2 & 73.4 & 14.1 & 5.6 & 43.5 & 13.4 & 0.7 & \textbf{48.1} & 91.2 & 16.4 & 54.1 & 18.8 & 79.2 & 70.6 & 75.7 & 36.8~
\\\midrule
\textbf{ConDA~(Ours)}~ & 11.8 & \textbf{9.0} & \textbf{49.1} & \textbf{76.3} & 7.2 & \textbf{18.0} & \textbf{62.6} & \textbf{15.2} & 0.0 & 47.9 & \textbf{92.3} & \textbf{34.9} & 59.4 & \textbf{27.9} & \textbf{83.2} & \textbf{82.3} & \textbf{79.3} & \textbf{42.3}~
\\\bottomrule
\end{tabular}}
\vspace{-0.32cm}
\label{tab:sg2bs}
\end{table*}

\noindent\textbf{Q4: \textit{How is domain concatenation superior to others?}}

\noindent\textbf{A4:} We compare five popular mixing techniques in Tab.~\ref{tab:concat-examples}. SimROD~\cite{SimROD} stitched samples from both domains as inputs for adaptation while MixUp~\cite{MixUp}, CutMix~\cite{CutMix}, CutOut~\cite{CutOut}, and Mix3D~\cite{Mix3D} are general regularization methods adopted for fully- and semi-supervised learning. It can be seen that they have shown sub-par performance in the RV representation for UDA in LiDAR segmentation. Differently, our approach is able to effectively leverage the spatial context of RV and combine both domains into an intermediate domain for fine-grained interactive learning and regularization.

\subsection{Comparison to the State of the Art}
\noindent\textbf{Benchmarking Results}. We compare ConDA with eight state-of-the-art methods on the Boston $\rightarrow$ Singapore and Singapore $\rightarrow$ Boston scenarios in Tab. \ref{tab:bs2sg} and Tab. \ref{tab:sg2bs}. In both cases, ConDA substantially outperforms other competitors in terms of mIoU and FIoU. Notably, in contrast to prior approaches that tend to improve performance on relatively easier classes (\textit{i.e.}, classes on which the source-only has already performed well), ConDA yields considerable performance gains on almost all classes. This strongly supports our findings that fine-grained objects/background interactions in-between domains are conducive for closing the domain gap and thus improving the adaptations.

\noindent\textbf{Qualitative Assessment}. Fig.~\ref{fig:qualitative} presents visualization results for different methods, \textit{i.e.}, self-training (CBST \cite{CBST}), adversarial training (DADA \cite{DADA}), and both (IntraDA \cite{IntraDA}). We observe that while the prior arts only give limited gains in certain areas, ConDA mitigates the false predictions holistically in most regions around the ego-vehicle. We accredit this to both the generalization ability provided by domain concatenation and the regularization enhancement offered by our anti-aliasing regularizer and entropy aggregator.

\noindent\textbf{Training Complexity}.
Fig.~\ref{fig:train_time} shows the training time of different methods. Note that the pseudo-label generation time is excluded since this operation is conventionally conducted offline \cite{xMUDA,CBST,BDL}. We find that while the inference speeds for these methods are similar (since they are sharing the same range-view segmentation backbone), ConDA is still faster than most adversarial methods \cite{AdaptSegNet,ADVENT,DADA} which rely on additional discriminators for learning domain-invariant features during the adaptation. Our work is comparable with MinEnt~\cite{ADVENT} and CBST~\cite{CBST} in terms of speed but provides much better adaptation performance.
\section{Conclusion}
We presented ConDA, a concatenation-based UDA framework that leverages the spatial coherency in-between the source and target domains for fine-grained interactive learning. 
Extensive experiments showed that ConDA can substantially improve the segmentation performance over baselines and competitive approaches. The robustness of our framework has shed light on its utility and potential for flexible deployment in the autonomous driving perception system. 

\small{\noindent\textbf{Acknowledgements}. This research is part of the programme DesCartes and is supported by the National Research Foundation, Prime Minister’s Office, Singapore under its Campus for Research Excellence and Technological Enterprise (CREATE) programme. This work is affiliated with the WP4 of the DesCartes programme, with an identity number: A-8000237-00-00.}

\clearpage
\clearpage

{\small
\bibliographystyle{ieeetr}
\bibliography{egbib}}

\end{document}